%% file: main.tex
\documentclass{article}

\PassOptionsToPackage{numbers, sort&compress}{natbib}

\usepackage[position,final]{support_files/neurips_2026}

\input{support_files/setup_script}

\Crefname{prop}{Proposition}{Propositions}
\crefname{prop}{Proposition}{Propositions}
\title{Position: Adopt Constraints Over Fixed Penalties \\in Deep Learning}

\author{%
Juan Ramirez\thanks{Correspondence to: \{\texttt{juan.ramirez}, \texttt{merajhse}\}@mila.quebec} \qquad
Meraj Hashemizadeh \qquad
Simon Lacoste-Julien$^{\dagger}$\\[1ex]
Mila - Quebec AI Institute and DIRO, Université de Montréal \\
$^{\dagger}$Canada CIFAR AI Chair
}

\begin{document}

\maketitle

\vspace{-1ex}
\begin{abstract}
    Deep learning models are increasingly subject to explicit requirements, or \emph{constraints}, such as safety levels, fairness limits, and sparsity budgets. In practice, such requirements are often handled through \emph{fixed weighted-sum penalization}: the constraints are added to the task loss with fixed coefficients, and the resulting scalarized objective is minimized. This position paper argues that fixed penalization is often ill-suited to deep learning problems with non-negotiable requirements, for three main reasons. First, in non-convex settings, the penalized and constrained problems are not, in general, equivalent, so a minimizer of the penalized objective need not solve the constrained problem. Second, a penalty coefficient sets an objective--constraint trade-off rather than the constraint level, so finding an acceptable coefficient often takes trial and error, each trial a full training run. Third, fixed penalization weakens hard requirements into soft penalties to be traded off against task performance. As a result, compliance is no longer built into the problem, and requirements can be quietly relaxed. We therefore argue that, when a deep learning problem specifies non-negotiable requirements, one should solve the constrained problem itself instead of a surrogate defined by fixed penalization.
\end{abstract}

\section{Introduction}

As deep learning systems are increasingly deployed in high-stakes applications, a growing number of requirements are imposed on them~\citep{montreal2018,madiega2021artificial}, often as externally specified thresholds such as acceptable safety levels~\citep{dai2024safe}, fairness limits~\citep{cotter2019proxy}, and compute or sparsity budgets~\citep{gallego2022controlled}. \emph{Constrained learning} encodes such requirements directly into training, optimizing performance subject to them. When a requirement is non-negotiable, constrained formulations build compliance into the problem specification itself, which makes success auditable.

In practice, constraints in deep learning are often handled through \emph{fixed weighted-sum penalization}, which adds them to the task loss with fixed coefficients and minimizes the resulting objective~\citep{raissi2019physics,louizos2017learning,zhang2019theoretically,chuang2021fair}; see \S\ref{sec:penalized}. Fixed penalization is easy to implement, requires little beyond differentiability, and fits naturally into standard unconstrained training pipelines.

This ease of use, however, often comes at a substantial cost. Our position is as follows:
\begin{center}
\begin{minipage}{\linewidth}\centering
\textit{When a deep learning problem specifies non-negotiable requirements, fixed weighted-sum penalization is often the wrong default. In such cases, one should start from the constrained formulation, and then choose a constrained method based on the problem’s structure and scale.}
\end{minipage}
\end{center}
This position rests on three limitations of fixed penalization.

First, in non-convex settings, there may be no penalty coefficient for which solving the penalized problem recovers the constrained solution (\S\ref{sec:exploration}, \cref{prop:concave}). In physics-informed neural networks, for instance, the governing equations must hold, yet they routinely enter the loss with fixed weights~\citep{raissi2019physics}. This practice typically yields either unusable solutions that violate the equations or solutions that satisfy them only at a severe cost in performance~\citep{lu2021physics}.

Second, choosing penalty coefficients is not an ordinary hyperparameter-tuning problem. Because the coefficient sets the objective--constraint trade-off instead of the constraint level, a poor choice means solving the wrong problem altogether. Finding an acceptable coefficient therefore often takes repeated trial and error, each trial costing a full training run (\S\ref{sec:tuning}, \cref{fig:sparsity}). A model built for on-device deployment, for instance, does not ship if it exceeds its memory or latency budget, yet meeting the budget via fixed penalization often requires retraining with one coefficient after another until the model fits~\citep{louizos2017learning,gallego2022controlled}.

Third, more fundamentally, fixed penalization weakens hard requirements into soft penalty terms that can be traded off against task performance (\S\ref{sec:drawbacks_accountability}). It replaces the constraint level, whose increase visibly relaxes the requirement, with a penalty coefficient, whose decrease relaxes it just as surely yet can pass for hyperparameter tuning. For instance, a language model required to produce no unsafe responses may then end up producing a few for a modest gain in helpfulness, a trade the formulation never states. Fixed penalization thus weakens accountability by construction.

We therefore advocate tailored constrained optimization methods, which target the constrained problem directly rather than a penalized surrogate. For deep learning, we particularly recommend gradient-based constrained methods, since deep learning problems are typically large-scale and non-convex and usually lack the structure that many classical constrained solvers require.

To illustrate how constrained methods address the limitations of fixed penalization, we use one of the simplest gradient-based constrained methods, the Lagrangian approach. This approach amounts to penalization whose coefficients are optimized jointly with the model rather than fixed in advance, which makes it directly comparable to fixed penalization. In \S\ref{sec:alternatives} we show that it recovers the constrained solution on a non-convex problem where fixed penalization cannot (\cref{table:concave}). We also show that it is much easier to tune: on a probability-constrained classification task, it lands on the constraint level across dual step-sizes spanning orders of magnitude, whereas no penalty coefficient on the same grid of values does (\cref{fig:robustness}). Finally, we argue that it keeps the convenience of fixed penalization, since a Lagrangian run can use the same standard deep learning machinery at only slightly higher computational cost (\cref{table:runtime}).

\textbf{Scope.}
We do not argue that every learning problem should be cast as constrained. When a quantity is a soft preference, a regularizer, or one objective among several, penalties or multi-objective formulations may be more appropriate. Nor do we claim that constrained formulations come free. Constrained problems can be hard to solve, and even when solved, a constraint satisfied on the training data need not hold on new data (see \S\ref{sec:challenges} for detailed counterarguments to our position). Classical penalty methods~\citep[\S17.1]{nocedal2006numerical}, which penalize constraint violations rather than the constraint functions themselves, are principled constrained optimization methods and fall outside our critique. Our critique begins once a requirement is non-negotiable and is handled through fixed weighted-sum penalization, as defined in \S\ref{sec:penalized}. The Lagrangian approach serves only as an illustration. The appropriate constrained method depends on the problem's structure and scale (\S\ref{sec:practical_takeaways}).

\textbf{Related Work and Contribution.}
The observations our argument rests on are known but scattered: weighted-sum scalarization is classically known to miss optimal trade-offs in non-convex multi-objective optimization \citep[\S3]{ehrgott2005multicriteria}, fixed penalties have long been criticized in deep learning~\citep{platt1988constrained,blog_1,gallego2024thesis}, and their tuning burden has motivated constrained methods in a range of applications~\citep{cotter2019proxy,gallego2022controlled,dai2024safe,roy2022direct}. Our contribution is to combine them into one argument and to distill from it a practical decision framework for choosing between penalties and constraints, and among constrained optimization methods (\S\ref{sec:practical_takeaways}).

\section{Fixed Weighted-Sum Penalization}
\label{sec:penalized}

Let $f: \X \rightarrow \reals$, $\vg: \X \rightarrow \reals^m$, and $\vh: \X \rightarrow \reals^n$ be differentiable functions over $\X \subseteq \reals^d$. We consider the constrained optimization problem
\begin{equation}
    \label{eq:const}
    \min_{\vx \in \X} \, f(\vx)
    \quad
    \text{s.t.}
    \quad
    \vg(\vx) \vleq \vepsilon_{\vg}
    \quad
    \text{and}
    \quad
    \vh(\vx) = \vepsilon_{\vh},
\end{equation}
where $\vepsilon_{\vg}$ and $\vepsilon_{\vh}$ are the constraint levels, $\vleq$ denotes element-wise inequality, and $\vg$ and $\vh$ are the inequality and equality constraints, respectively. A point $\vx \in \X$ is \textit{feasible} if it satisfies the constraints, and a feasible point $\xstar$ is a \textit{constrained minimizer} if $f(\xstar) \leq f(\vx)$ for all feasible $\vx$.

In machine learning, $f$ typically captures task performance, while the constraints encode requirements such as sparsity~\citep{gallego2022controlled}, fairness~\citep{cotter2019proxy}, or safety~\citep{dai2024safe}. The key distinction from multi-objective optimization is that here $\vg$ and $\vh$ are not quantities to be optimized but requirements to be \emph{satisfied}. Solutions that violate them are inadmissible, regardless of their objective value.\footnote{In practice, small numerical violations of the constraints are often tolerated.}

A common way to tackle \cref{eq:const} in deep learning is to drop the hard constraints and instead minimize a fixed weighted sum of the objective and constraint functions \citep[\S3]{ehrgott2005multicriteria}:
\begin{equation}
    \label{eq:penalized}
    \min_{\vx \in \X}
    \P(\vx)
    \defas
    f(\vx) + \vcg^\top \vg(\vx) + \vch^\top \vh(\vx),
\end{equation}
where $\vcg \vgeq \vzero$ and $\vch$ are fixed coefficients. We call
\cref{eq:penalized} \textit{fixed weighted-sum penalization}, or fixed
penalization for short.

Fixed penalization is popular in deep learning for three reasons:
\blobletter{1} it yields a single unconstrained objective that standard training pipelines can minimize directly, requiring no projections, feasibility-maintaining updates, or extra optimization variables;
\blobletter{2} it applies to any differentiable problem and requires no other structural assumption; and
\blobletter{3} its per-step cost is close to that of unconstrained training, since the gradient of $\P$ can be computed in a single backward pass.

Fixed penalization can also be principled for convex problems:

\begin{prop}[Convex justification for fixed weighted-sum penalization]
    \label{prop:penalized}
    Let $f$ and $\vg$ be differentiable and convex, let $\vh$ be affine, and assume $\X$ is convex. Let $\xstar \in \interior(\X)$ be a constrained minimizer of \cref{eq:const}, and suppose there exist Lagrange multipliers $\lambdastar \vgeq \vzero$ and $\mustar$ at $\xstar$. Then
    \begin{equation}
        \xstar
        \in
        \arg\min_{\vx \in \X}
        \Big[
        f(\vx) + (\lambdastar)^\top \vg(\vx) + (\mustar)^\top \vh(\vx)
        \Big].
    \end{equation}
    Proof. \emph{See \hyperlink{proof:penalized}{Proof of Proposition \ref{prop:penalized}} in Appendix~\ref{app:proofs}.}
\end{prop}

Thus, in the convex setting, fixed penalization is justified \emph{provided the correct coefficients are known}. With $\vcg=\lambdastar$ and $\vch=\mustar$, the constrained minimizer is a minimizer of the weighted sum, and the unique one when $f$ is strictly convex. More broadly, under convexity, varying the coefficients lets fixed penalization recover the Pareto-optimal trade-offs between the objective and constraint functions \citep[Thm.~3.1.4]{miettinen1999nonlinear}.

However, this justification requires convexity, which deep learning rarely offers, and the correct coefficients, which are rarely known.

\section{Limitations of Fixed Penalization}
\label{sec:drawbacks}

In this section, we highlight three limitations of fixed penalization in constrained deep learning: in non-convex problems it is not, in general, equivalent to the constrained problem and may fail to recover a constrained minimizer; its coefficients are costly to tune; and it weakens hard requirements into soft trade-offs.

\subsection{The Penalized and Constrained Problems Are \textit{Not} Equivalent}
\label{sec:exploration}

The following concave example, adapted from \citet{blog_1,blog_2}, shows that there may be no choice of fixed coefficients for which minimizing \cref{eq:penalized} recovers a constrained minimizer of \cref{eq:const}. Instead, each coefficient yields either an infeasible solution or one that over-satisfies the constraint at a cost in the objective (\cref{fig:pareto}, Appendix~\ref{app:concave}).

\begin{example}[Concave 1D problem]
    \label{ex:concave}
    \normalfont
    Consider the following constrained optimization problem:
    \begin{equation}
        \min_{x \in [0, \pi/2]} \, f(x) \defas \cos(x)
        \qquad
        \text{s.t.} \quad g(x) \defas \sin(x) \leq \epsilon,
    \end{equation}
    where $\epsilon \in (0,1)$.
    This problem violates the convexity hypothesis of \cref{prop:penalized}, since both $f$ and $g$ are concave on $[0,\pi/2]$.
\end{example}

\begin{prop}[Solution to \cref{ex:concave}]
    \label{prop:solution}
    The constrained minimizer of \cref{ex:concave} is $x^*=\arcsin(\epsilon)$, with Lagrange multiplier $\lambda^*=\epsilon/\sqrt{1-\epsilon^2}$. At this point, $g(x^*)=\epsilon$ and $f(x^*)=\sqrt{1-\epsilon^2}$, so the constraint is active.
    Proof. \emph{See \hyperlink{proof:solution}{Proof of Proposition \ref{prop:solution}} in Appendix~\ref{app:proofs}.}
\end{prop}

\begin{prop}[The failure of fixed penalization]
    \label{prop:concave}
    Consider the penalized formulation of \cref{ex:concave}:
    \begin{equation}
        \label{eq:concave_penalized}
        \min_{x \in [0, \pi/2]} \, \P_c(x) \defas \cos(x) + c \, \sin(x),
    \end{equation}
    where $c \geq 0$ is a penalty coefficient. Then 
    \blobletter{1} for every $c \in [0,1)$, the unique minimizer of \cref{eq:concave_penalized} is $x_c=\pi/2$, with $f(\pi/2)=0$ and $g(\pi/2)=1>\epsilon$;
    \blobletter{2} for every $c>1$, the unique minimizer is $x_c=0$, with $f(0)=1$ and $g(0)=0$;
    and \blobletter{3} when $c=1$, both points are minimizers: $x_c \in \{0, \pi/2\}$.
    Proof. \emph{See \hyperlink{proof:concave}{Proof of Proposition \ref{prop:concave}} in Appendix~\ref{app:proofs}.}
\end{prop}

\Cref{prop:concave} shows that no choice of penalty coefficient recovers the constrained minimizer of \cref{ex:concave}. For $c<1$, the penalized formulation returns the minimizer of $f$ alone, $x_c=\pi/2$, which ignores the constraint and is infeasible since $g(\pi/2)=1>\epsilon$. For $c>1$, it returns the minimizer of $g$ alone, $x_c=0$, which over-satisfies the constraint ($g(0)=0<\epsilon$) at a substantial cost in the objective. Setting $c$ to the Lagrange multiplier $\lambda^*=\epsilon/\sqrt{1-\epsilon^2}$ does not help either: the minimizer of $\P_{\lambda^*}$ is still $\pi/2$ or $0$, never $x^*=\arcsin(\epsilon)$.

This failure arises because, for every coefficient choice, the penalized objective $\P_c(x)=\cos(x)+c\sin(x)$ is concave on $[0,\pi/2]$, so its minimizers lie at the domain boundaries, whereas the constrained minimizer lies in the interior. More generally, in non-convex problems, fixed penalization can reach only the points of the objective--constraint trade-off curve that lie on its convex envelope~\citep[\S3]{ehrgott2005multicriteria}. Wherever this curve is locally non-convex at the constraint level, the constrained minimizer is unreachable by any fixed weighted sum of the objective and constraint functions.\footnote{The constrained formulation, known in multi-objective optimization as the $\epsilon$-constraint method, reaches every Pareto-optimal trade-off, including those a weighted sum misses~\citep[\S4.1]{ehrgott2005multicriteria}.} Checking for such non-convexity, however, would require knowing the trade-off curve around the constraint level, which means solving the constrained problem in the first place.

\subsection{Penalty Coefficients Are Costly to Tune}
\label{sec:tuning}

Even under convexity, fixed penalization recovers a satisfactory solution only if its coefficients $\vcg$ and $\vch$ are close to the Lagrange multipliers $\lambdastar$ and $\mustar$ at the constraint levels $(\vepsg,\vepsh)$. The difficulty is that the coefficients do not directly encode these levels, but instead define a trade-off between the objective and the constraint functions. As a result, a poor coefficient does not merely mean solving the right problem poorly, but rather solving the wrong problem altogether.

\begin{wrapfigure}{r}{0.36\textwidth}
    \vspace{-2.5ex}
    \centering
    \includegraphics[width=0.93\linewidth]{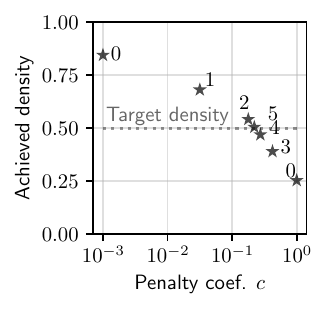}
    \vspace{-2ex}
    \caption{
    Solving a sparsity-constrained classification problem. Starting from coefficients yielding low (25\%) and high (85\%) model density, \textbf{five bisection search steps are needed to reach the desired 50\% density}. \captioncomment{Numbers mark iterations (endpoints are ``0''). Replicates Fig.~5(b) of \citet{gallego2022controlled} (Appendix~\ref{app:sparsity}).}}
    \label{fig:sparsity}
    \vspace{-2ex}
\end{wrapfigure}

Because the appropriate trade-off is unknown \emph{a priori}, practitioners often identify satisfactory coefficients only through repeated trial and error, where each trial is a full training run~\citep{gallego2022controlled}. \Cref{fig:sparsity} illustrates this burden on a neural network trained under a constraint requiring at least $50\%$ parameter sparsity. With bisection search, the penalized problem must be solved seven times (two bracketing runs and five bisection steps) before arriving at a satisfactory solution, one that meets the constraint to within one percentage point but is not so sparse that performance is unnecessarily degraded. A constrained optimization method, by contrast, can solve the same problem in a single run~\citep{gallego2022controlled}.

Several factors make tuning penalty coefficients difficult. First, a coefficient has no meaning on its own. It sets only the weight of a constraint function relative to the objective and other constraints, which may all have different units, so there is no straightforward way to initialize it. Second, the relationship between a coefficient and the resulting objective and constraint values is often nonlinear, so the effect of adjusting it is hard to predict even within a single task. Third, coefficients rarely transfer across problems: their effects depend on the scale and landscape of the objective and constraint functions, so a value that works for one architecture or task may fail on another~\citep{robey2021adversarial}. Finally, these difficulties compound with multiple constraints, since the coefficients must be chosen jointly.

By contrast, the constraint levels $(\vepsg,\vepsh)$ in constrained formulations directly specify the desired outcome and share the units of the constraint functions. As problem specifications rather than coefficients of a surrogate problem, they transfer more naturally across architectures and tasks.

\subsection{Fixed Penalization Weakens Hard Requirements into Soft Trade-offs}
\label{sec:drawbacks_accountability}

Even setting aside the issues above, fixed penalization changes the meaning of the problem, which no longer asks the model to meet the non-negotiable requirement. Whereas the constrained problem states a constraint level, the penalized problem drops it, leaving only a penalty term to be traded off against task performance. Training thus commits the model to no particular level; the one it ends up reaching is a byproduct of the chosen coefficient.

Without a stated level, the requirement is easy to relax, deprioritize, or leave unreported, which weakens accountability by construction. Relaxing it amounts to lowering a penalty coefficient, an adjustment that passes for hyperparameter tuning. For instance, a classifier required to keep a fairness gap at zero can be allowed a small gap in exchange for accuracy, and then be evaluated by accuracy alone, with the fairness gap it actually reaches never stated.

A constraint, by contrast, keeps the requirement accountable by stating the level the model must meet~\citep{cotter2019proxy,roy2022direct,gallego2022controlled}. The stated level is a claim about the model that is true or false, so it can be checked, reported, and contested, and relaxing it is a visible change to the problem. Under fixed penalization, the level reached can still be measured, but there is no stated level to check it against.

\section{Alternatives to Fixed Penalization}
\label{sec:alternatives}

The three limitations of \S\ref{sec:drawbacks} come from solving a fixed penalized surrogate rather than the constrained problem itself, so the alternative is to target it directly. Some constrained methods do so by exploiting problem structure to enforce feasibility, for example through projection, reparameterization, or specialized architectures. In modern non-convex deep learning, however, such structure is often unavailable.

We therefore use the Lagrangian approach as our illustration, since it applies to the same broad class of differentiable problems as fixed penalization. Lagrangian methods are supported in deep learning frameworks through libraries such as Cooper~\citep{gallegoPosada2025cooper} and TFCO~\citep{cotter2019tfco}, and are already used across many constrained deep learning problems~\citep{cotter2019proxy,robey2021adversarial,stooke2020responsive,elenter2022lagrangian,hounie2023automatic,ramirez2025feasible}, including at scale: sparsity on ImageNet~\citep{gallego2022controlled}, fairness across 100 sub-groups~\citep{hashemizadeh2024balancing}, and safety thresholds in language model alignment~\citep{dai2024safe}. The Lagrangian approach is also the constrained method closest to fixed penalization: the fixed coefficients become multipliers that adapt during training.

\subsection{The Basic Lagrangian Mechanism}
\label{sec:lagrangian}

The Lagrangian approach recasts \cref{eq:const} as a min-max problem over the Lagrangian $\Lag$:
\begin{equation}
    \label{eq:lag_min_max}
    \min_{\vx \in \X}
    \max_{\vlambda \vgeq \vzero,\vmu}
    \Lag(\vx,\vlambda,\vmu)
    \defas
    f(\vx)
    + \vlambda^\top \big[\vg(\vx)-\vepsilon_{\vg}\big]
    + \vmu^\top \big[\vh(\vx)-\vepsilon_{\vh}\big],
\end{equation}
where $\vlambda \vgeq \vzero$ and $\vmu$ are the multipliers associated with the inequality and equality constraints, respectively. We refer to $\vx$ as the primal variables and to $\vlambda$ and $\vmu$ as the dual variables. Maximizing over the multipliers enforces the constraints, making \cref{eq:lag_min_max} equivalent to the constrained problem \cref{eq:const}.

A simple first-order algorithm for solving \cref{eq:lag_min_max} is alternating gradient descent--ascent:
\begin{align}
    \label{eq:gda_updates}
    \begin{gathered}
        \begin{aligned}
            \vmu_{t+1}
            &\leftarrow
            \vmu_t + \lrd \big[\vh(\vx_t)-\vepsilon_{\vh}\big],
            \qquad
            \vlambda_{t+1}
            &\leftarrow
            \Big[\vlambda_t + \lrd \big[\vg(\vx_t)-\vepsilon_{\vg}\big]\Big]_+
        \end{aligned} \\
        \vx_{t+1}
        \leftarrow
        \Big[\vx_t - \lrp \nabla_{\vx}\Lag(\vx_t,\vlambda_{t+1},\vmu_{t+1})\Big]_{\X},
    \end{gathered}
\end{align}
where $[\cdot]_+$ and $[\cdot]_{\X}$ denote projection onto $\reals_{\vgeq \vzero}^m$ and $\X$, respectively, and $\eta_{\{\text{primal},\text{dual}\}}$ are step-sizes.

The primal update in \cref{eq:gda_updates} coincides with a gradient descent step on the penalized objective of \cref{eq:penalized}, with $\vc_{\vg}=\vlambda_{t+1}$ and $\vc_{\vh}=\vmu_{t+1}$. Fixed penalization is the special case of \cref{eq:gda_updates} with $\lrd=0$, in which the multipliers never leave their initial values $\vc_{\vg}=\vlambda_0$ and $\vc_{\vh}=\vmu_0$.

With $\lrd>0$, the multipliers, typically initialized at zero,\footnote{Initializing the multipliers at zero presumes no objective--constraint trade-off in advance. Any other initial value is a hand-chosen coefficient, as arbitrary as a penalty coefficient (\S\ref{sec:tuning}), though unlike a penalty coefficient it can be corrected by the dynamics.} move in response to the constraints. An inequality-constraint multiplier increases when its constraint is violated, placing more weight on reducing the violation; when the constraint is satisfied with slack, it decreases, letting the primal updates focus on the objective.\footnote{An equality multiplier behaves analogously, increasing when its constraint function exceeds the target and decreasing when it falls short.} The appropriate trade-off between minimizing the objective and satisfying the constraints is thus learned during optimization rather than chosen by hand in advance.

\subsection{How Tailored Constrained Methods Address the Limitations of Fixed Penalization}
\label{sec:contrasting}

\textbf{Constrained methods can recover solutions that fixed penalization misses.}
On the non-convex problem of \cref{ex:concave}, the Lagrangian approach recovers the constrained minimizer $x^*=\arcsin(\epsilon)$ at every constraint level tested (\cref{table:concave}), provided the multipliers are updated by a PI controller with a sufficiently large damping coefficient~\citep{stooke2020responsive,sohrabi2024nupi}.\footnote{Plain gradient descent--ascent on the Lagrangian does not converge on this problem~\citep{platt1988constrained}, because the Lagrangian, like the penalized objective, is concave in $x$ for every fixed multiplier (Appendix~\ref{app:concave}). A PI controller with a sufficiently large damping coefficient, or equivalently the Augmented Lagrangian method~\citep{ramirez2025dual}, stabilizes these dynamics. Above a threshold, every damping coefficient allows local convergence to any regular constrained minimizer~\citep[Thm.~3]{ramirez2025dual}.} By contrast, fixed penalization cannot recover $x^*$ for any coefficient (\cref{prop:concave}).

\input{tables/concave_table}

\begin{figure*}[t]
    \centering
    \vspace{-2ex}
    \includegraphics[width=\linewidth]{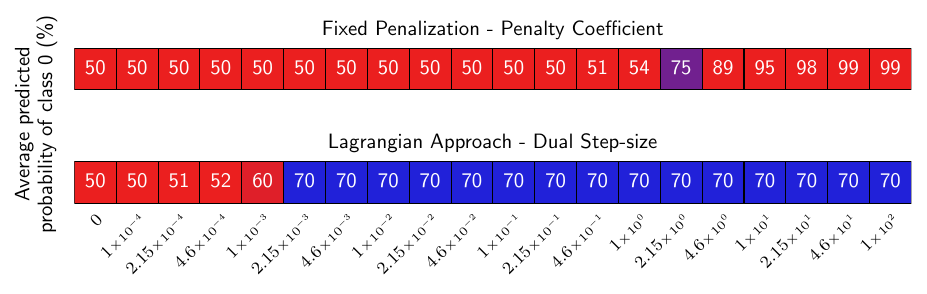}
    \vspace{-4ex}
    \caption{Solving a binary classification task under a constraint that the average predicted probability of class ``0'' be at least 70\%. \textbf{No penalty coefficient in the sweep lands on the 70\% target. Each either leaves the constraint violated or over-satisfies it at a cost in accuracy, whereas the Lagrangian approach lands on the target across a broad range of dual step-sizes.} Cells are blue within one percentage point of the target and shade to red at eleven or more points away. \captioncomment{Data and model from Figure~2 of \citet{cotter2019proxy} (details in Appendix~\ref{app:rate}).}}
    \label{fig:robustness}
    \vspace{-1ex}
\end{figure*}

\textbf{A dual step-size is easier to tune than a penalty coefficient.}
Because the constrained problem is defined by its constraint levels, the dual step-size, like any other optimization hyperparameter, does not affect \emph{which} problem is solved, only \emph{how well}. A penalty coefficient, by contrast, determines which problem is solved. Moreover, as for any step-size, a wide range of values often works (\cref{fig:robustness}), whereas a coefficient that recovers the constrained minimizer, if one exists, is often a single value rather than a range. Choosing a dual step-size poorly thus means solving the right problem poorly; choosing a penalty coefficient poorly means solving the wrong problem \mbox{altogether}.

\Cref{fig:robustness} illustrates both points on a binary classification task adapted from \citet{cotter2019proxy}, where the average predicted probability of class ``0'' must be at least 70\%. No penalty coefficient in the sweep, which spans six orders of magnitude, lands on the 70\% target.\footnote{A finer sweep may well find a penalty coefficient between $c=1$ and $c=2.15$ that lands on the target. The point is rather that the same grid of 20 values contains 15 suitable dual step-sizes and no suitable penalty coefficient.} The 14 smallest leave the average probability near 50\%, reaching only 54\% at $c=1$, and the six largest push it to 75\% or more. The Lagrangian approach, by contrast, lands on the target for 15 of the 20 dual step-sizes on the same grid, missing only for the five smallest, for which the multiplier grows too slowly to converge within the training budget. At the constraint level, the Lagrangian approach reaches 84\% accuracy, whereas the penalty coefficient closest to the target yields a feasible solution with only 74\%.

\textbf{Feasibility is pursued during training, not only checked after it.}
A constrained method keeps the constraint level in the problem it solves, so the level guides training and remains the standard against which the result is checked. In the Lagrangian approach, the multiplier updates of \cref{eq:gda_updates} keep adjusting the weight on each constraint until it is satisfied.\footnote{A multiplier that never stops growing often signals that its constraint cannot be satisfied, which makes it a useful diagnostic in practice.} Fixed penalization, by contrast, may converge to a solution that still violates the constraints, with no stated level to check it against.

\textbf{Per-run overhead is small.}
The first-order Lagrangian updates of \cref{eq:gda_updates} cost little more than their penalized counterparts. The primal update requires the same forward and backward passes through the objective and constraint functions as fixed penalization. The multiplier updates add only a few scalar operations per constraint. For any moderate number of constraints, this cost is negligible next to those passes and shrinks further as models grow. \Cref{table:runtime} quantifies this overhead in two settings: about $5\%$ for probability-constrained classification and under $2\%$ for sparsity-constrained training.

The cost comparison that matters, however, is \emph{per solution} rather than per training run. Tuning a penalty coefficient means training repeatedly, whereas the Lagrangian approach can reach the target in a single run (\S\ref{sec:tuning}). Despite its extra machinery and per-run overhead, the Lagrangian approach can therefore be much cheaper in practice.

\begin{table}[h!]
    \centering
    \small
    \caption{Wall-clock comparison over five runs on an NVIDIA L40S GPU in two illustrative settings. Constrained training adds only a small per-run overhead.}
    \label{table:runtime}
    \begin{tabular}{lccc}
        \toprule
        Experiment & Penalized (median, std.) & Constrained (median, std.) & Overhead \\
        \midrule
        Probability constraint
        & 5.40 s (0.27 s)
        & 5.68 s (0.31 s)
        & +5.2\% \\
        Sparsity constraint
        & 8.17 min (0.19 min)
        & 8.30 min (0.07 min)
        & +1.6\% \\
        \bottomrule
    \end{tabular}
\end{table}

\section{Constraints Versus Fixed Penalties: Counterarguments}
\label{sec:challenges}

Of the five counterarguments below, the first questions whether a constraint is the right formulation at all; the other four accept the formulation but question whether it is worth adopting in practice.

\textbf{Penalties may better reflect the problem.}
A penalty is often the more natural abstraction when the quantity of interest is not a non-negotiable requirement, but rather something to discourage, control, or trade off against task performance. Examples include regularizers, soft preferences, and proxy quantities for which no clear threshold is available. In such cases, imposing a hard constraint may draw a sharper acceptable--unacceptable boundary than the problem itself warrants.

\textbf{Fixed penalization is practical.}
Fixed penalization remains attractive because it yields an unconstrained problem and so preserves standard model training pipelines. One defines a single scalar loss, uses familiar optimizers and schedules, and avoids introducing dual variables or other constrained-optimization machinery. This simplicity reduces engineering effort and removes the need to select, implement, and tune a constrained method for the problem at hand. Ease of use is not decisive, however, since gradient-based constrained methods like the Lagrangian approach can often preserve much of the same pipeline while targeting the requirement directly.

\textbf{Fixed penalization is often an adequate baseline.}
In many applications, fixed penalization is used not only because it is simple, but also because it often provides a serviceable baseline. Moderate coefficient tuning may suffice to obtain solutions with acceptable objective--constraint trade-offs. In such settings, practitioners may reasonably prefer a familiar and empirically adequate method to a more specialized constrained approach. However, when compliance is non-negotiable, an acceptable trade-off is not the same as meeting the requirement, and one should take the constrained route.

\textbf{Solving the right problem can still be hard.}
Even when the constrained formulation is the right one, solving it can still be difficult, especially in deep learning~\citep{elenter2024near}. In this setting, the constraints are often stochastic, so they must be estimated from mini-batches, and the problems are non-convex, so solvers lose their global convergence guarantees and may oscillate or even diverge. Many classical constrained methods do not transfer cleanly to such problems, and those that do transfer require more specialized tooling than unconstrained training. The Lagrangian approach, for instance, may need damping to converge, which brings extra hyperparameters~\citep{stooke2020responsive,sohrabi2024nupi}. Yet these difficulties are an argument for better constrained methods, not for replacing the constrained problem with a penalized surrogate that is just as stochastic and non-convex and only moves the difficulty into choosing the coefficient (\S\ref{sec:tuning}).

\textbf{Constraint satisfaction may not generalize.}
When constraints are data-dependent, satisfying them on the training data does not, in general, guarantee satisfaction on unseen examples~\citep{cotter2019generalization,hashemizadeh2024balancing}. Thus, a model may appear compliant during development while still violating the requirement at deployment, where compliance actually matters. But penalization does not remove this statistical difficulty, since requirements enforced via penalties need not generalize any better.

These considerations mark the limits of our position, but do not overturn it. Our disagreement is not with the claim that fixed penalties can be useful or empirically adequate, but with treating that convenience as a sufficient reason to soften a non-negotiable requirement into a fixed trade-off.

\section{How to Choose the Right Formulation and Method in Practice}
\label{sec:practical_takeaways}

We now turn to the practical question: when a learning problem involves a quantity that may encode a non-negotiable requirement, should one use a penalty or a constrained formulation, and if the latter, which constrained method? Rather than survey constrained methods exhaustively, we distill our position into a coarse decision framework for practitioners, which asks three questions in turn: whether the quantity is a requirement at all, whether the feasible set has exploitable structure, and at what scale the problem arises.

\textbf{First decide whether the quantity is a preference or a requirement.}
Whether to use a constraint is a conceptual choice, not an algorithmic one. If the quantity of interest is a soft preference, a regularizer, or one objective among several, a penalized or multi-objective formulation may be more appropriate. If instead it encodes a non-negotiable requirement---something that must remain above, below, or within a meaningful threshold---then it should be modeled as a constraint. One should not adopt constraints merely because constrained optimization is available, but neither should one weaken a non-negotiable requirement into a penalty term merely because unconstrained optimization is easier.

\textbf{When the feasible set has exploitable structure, enforce feasibility directly.}
Methods that enforce feasibility directly include projected methods when projections are cheap~\citep{levitin1966constrained,goldstein1964convex,bertsekas1976goldstein}, reparameterizations when the feasible set admits an unconstrained parameterization~\citep{geotorch}, constrained output layers when the constraint applies to the model's outputs~\citep{min2024hardnet}, Riemannian methods when the feasible set is a smooth manifold~\citep{absil2008optimization,boumal2023introduction}, and feasible-direction methods when feasible directions are cheap to compute, as over a convex polytope~\citep{frank1956algorithm,zoutendijk1960methods}. In all these cases, once training starts from a feasible point, every iterate remains feasible, so the requirement holds by construction throughout training.

\textbf{Outside the large-scale deep learning regime, consider richer constrained solvers.}
When the problem offers no exploitable structure, a solver can impose structure locally instead, by solving a sequence of tractable approximations to it. The classic example is sequential quadratic programming~\citep{wilson1963simplicial,han1977globally,powell1978convergence}, which solves at each iteration a quadratic model of the Lagrangian under linearized constraints and typically converges in far fewer iterations than gradient-based methods~\citep[Ch.~18]{nocedal2006numerical}. Each iteration, however, needs the constraint Jacobians and the solution of a subproblem, both costly at the scale of modern deep learning. Recent variants extend it to stochastic, possibly non-convex objectives under deterministic constraints~\citep{berahas2021sequential,curtis2024sequential}, but their iterations are just as costly.

\textbf{In large-scale deep learning without exploitable structure, start from gradient-based constrained methods.}
Gradient-based constrained methods need only (sub)gradients of the objective and constraints, combined through multipliers or penalty coefficients that adapt during training, so they add little overhead to unconstrained training and carry over to stochastic objectives and constraints. While we do not argue that gradient-based methods are the only ones that scale, these properties make them especially well matched to large-scale deep learning.

The Lagrangian approach, one of the simplest of these methods, is a reasonable first choice. Plain gradient descent--ascent often suffices, as on the classification task of \cref{fig:robustness}. When it oscillates or fails to converge, as in \cref{ex:concave}, the natural next step is to damp the multiplier updates with PI control~\citep{stooke2020responsive,sohrabi2024nupi}, or equivalently the Augmented Lagrangian method~\citep{hestenes1969multiplier,powell1969method,rockafellar1973dual,ramirez2025dual}.

Although intentionally coarse, this framework is a useful starting point for what to do instead of defaulting to penalties, one that application-specific judgment should then refine.

\textbf{In summary}: use penalties for soft preferences and constraints for non-negotiable requirements. If the feasible set has exploitable structure, enforce feasibility directly. If not, the next question is scale: outside large-scale deep learning, richer constrained solvers may be preferable; in large-scale deep learning, gradient-based constrained methods, like the Lagrangian approach, are the natural starting point.

\section{Conclusion}

When deep learning problems involve non-negotiable requirements, fixed penalization is often the wrong default. In non-convex settings, the penalized and constrained problems need not be equivalent; choosing penalty coefficients is costly; and even when both are set aside, fixed penalization turns hard requirements into soft penalties to be traded off against task performance, which weakens accountability. One should therefore solve the constrained problem itself, with a method suited to its structure and scale.

We do not claim that every learning problem should be cast as constrained, nor that constrained methods are free of practical limitations. Constrained optimization can be more delicate than unconstrained training, and its tooling less routine. These costs, however, come from solving the right problem rather than a more convenient surrogate, and they are worth paying when the requirement is non-negotiable.

\textbf{Call for Research: Toward Seamless Constrained Training.}
If gradient-based constrained methods are to become as routine as unconstrained training, the central problem on the optimization side is robustness: they must work essentially out of the box, across a wide range of tasks and with minimal hyperparameter tuning, much as Adam does for unconstrained minimization~\citep{kingma2015adam}. Today these methods often fall short, for two reasons. The primal-dual dynamics may oscillate or even diverge (\S\ref{sec:challenges}), and when the constraints are estimated from mini-batches, the multiplier updates follow a noisy estimate of the violation, so the multipliers fluctuate rather than settle. Remedies for oscillation and noise bring hyperparameters of their own, however~\citep{stooke2020responsive,sohrabi2024nupi}, and a method that requires tuning new hyperparameters for every new problem, such as damping coefficients and, to a lesser extent, dual step-sizes, discourages adoption.

On the statistical side, constraint satisfaction can generalize beyond the training data~\citep{chamon2020probably}, but not always well enough~\citep{hashemizadeh2024balancing}. Yet a requirement imposed during training must hold at deployment, and few methods, among them that of \citet{cotter2019generalization}, target this gap directly.

Together, these problems define the work needed to make constrained learning as robust, usable, and reliable as unconstrained training. Whatever the state of the methods, a non-negotiable requirement is not a fixed penalty but a constraint.

\acksection

This work was supported by RBC
Borealis through the RBC Borealis AI Global Fellowship Award, by the Canada CIFAR AI Chair program (Mila), by the NSERC Discovery Grant RGPIN-2025-05123, by an unrestricted gift from Google, and by Samsung Electronics Co., Ltd. Simon Lacoste-Julien is a CIFAR Associate Fellow in the Learning in Machines \& Brains program.

We thank Jose Gallego-Posada and Lucas Maes for helpful discussions during the conceptualization of this work.

Many of the ideas presented in this work arose from discussions during the development of various research papers on constrained deep learning, especially \citet{gallego2022controlled}. We therefore thank our collaborators Yoshua Bengio, Juan Elenter, Akram Erraqabi, Golnoosh Farnadi, Jose Gallego-Posada, Ignacio Hounie, Alejandro Ribeiro, Motahareh Sohrabi, Rohan Sukumaran, and Tianyue (Helen) Zhang.

\bibliography{references.bib}
\bibliographystyle{abbrvnat} 

\newpage
\include{appendix}

\end{document}

%% file: support_files/setup_script.tex
\usepackage[utf8]{inputenc}
\usepackage[T1]{fontenc}

\usepackage[dvipsnames]{xcolor}
\definecolor{linkcolor}{RGB}{0,120,130}
\usepackage[colorlinks=true,
    linkcolor=linkcolor,
    citecolor=linkcolor,
    filecolor=linkcolor,
    urlcolor=linkcolor,
    pagebackref=true]{hyperref} 
\hypersetup{
    colorlinks=true,
    linkcolor=linkcolor,
    filecolor=magenta,      
    urlcolor=ForestGreen,
    }
\usepackage{url}

\renewcommand*\backref[1]{\ifx#1\relax \else (Cit. on p. #1) \fi}

\usepackage{todonotes}
\usepackage{algorithm}
\usepackage{amssymb}
\usepackage{amsfonts}
\usepackage{xfrac}
\usepackage{amsmath}
\usepackage{amsthm}
\usepackage{amssymb}
\usepackage{cancel}
\usepackage{mathtools}
\usepackage{bbm}
\usepackage{bm}
\usepackage{cases}
\usepackage{mathalpha}
\usepackage{microtype}
\usepackage{enumitem}
\usepackage{multirow}
\usepackage{booktabs}
\usepackage{tablefootnote}
\usepackage{adjustbox}
\usepackage{caption}
\usepackage{subcaption}
\usepackage{wrapfig}

\newtheorem{prop}{Proposition}
\newtheorem{example}{Example}

\usepackage{scrwfile}
\TOCclone[\appendixname]{toc}{atoc}
\newcommand\StartAppendixEntries{}
\AfterTOCHead[toc]{%
  \renewcommand\StartAppendixEntries{\value{tocdepth}=-10000\relax}%
}
\AfterTOCHead[atoc]{%
  \edef\maintocdepth{\the\value{tocdepth}}%
  \value{tocdepth}=-10000\relax%
  \renewcommand\StartAppendixEntries{\value{tocdepth}=\maintocdepth\relax}%
}
\newcommand*\appendixwithtoc{%
  \appendix
  \renewcommand{\baselinestretch}{1.3}\normalsize
  \addtocontents{toc}{\protect\StartAppendixEntries}
  \listofatoc
  \renewcommand{\baselinestretch}{1.0}\normalsize
}

\input{support_files/math_commands}

\usepackage[indLines=false, noEnd=false]{algpseudocodex}
\algrenewcommand\algorithmicrequire{\textbf{Input:}}

\definecolor{captiongray}{RGB}{80,80,80}
\newcommand{\captioncomment}[1]{{\color{captiongray} \footnotesize #1}}

\PassOptionsToPackage{numbers,sort}{natbib}

\usepackage{microtype}
\usepackage{graphicx}
\usepackage{booktabs}

\usepackage{hyperref}

\usepackage{amsmath}
\usepackage{amssymb}
\usepackage{mathtools}
\usepackage{amsthm}

\usepackage[capitalize,noabbrev]{cleveref}

\Crefname{theorem}{Theorem}{Theorems}%
\Crefname{prop}{Prop.}{Props.}
\Crefname{example}{Example}{Examples}

%% file: support_files/math_commands.tex
\usepackage{pifont}

\newcommand{\reals}{\mathbb{R}}

\newcommand{\defas}{\triangleq}

\newcommand{\vzero}{\boldsymbol{0}}

\newcommand{\vc}{\boldsymbol{c}}

\newcommand{\vg}{\boldsymbol{g}}
\newcommand{\vh}{\boldsymbol{h}}

\newcommand{\vw}{\boldsymbol{w}}
\newcommand{\vx}{\boldsymbol{x}}

\newcommand{\vz}{\boldsymbol{z}}

\newcommand{\vepsilon}{\boldsymbol{\epsilon}}
\newcommand{\vphi}{\boldsymbol{\phi}}

\newcommand{\Lag}{\mathcal{L}}

\newcommand{\vlambda}{\boldsymbol{\lambda}}
\newcommand{\vmu}{\boldsymbol{\mu}}

\newcommand{\xstar}{\vx^*}
\newcommand{\lambdastar}{\vlambda^*}
\newcommand{\mustar}{\vmu^*}

\renewcommand{\P}{\mathcal{P}}

\newcommand{\vepsg}{\vepsilon_{\vg}}
\newcommand{\vepsh}{\vepsilon_{\vh}}

\newcommand{\vcg}{\vc_{\vg}}
\newcommand{\vch}{\vc_{\vh}}

\newcommand{\lrp}{\eta_{\text{primal}}}
\newcommand{\lrd}{\eta_{\text{dual}}}

\newcommand{\X}{\mathcal{X}}

\newcommand{\vleq}{\preceq}
\newcommand{\vgeq}{\succeq}

\newcommand{\blobletter}[1]{\raisebox{.5pt}{\textcircled{\raisebox{-.7pt}{{\hspace{-0.8mm} \small #1}}}}}

\definecolor{lightgray}{RGB}{230, 230, 230}
\definecolor{mathred}{RGB}{204, 69, 90}
\definecolor{mathblue}{RGB}{4, 78, 112}
\definecolor{mathgreen}{RGB}{1, 135, 70}

\DeclareMathOperator*{\argmin}{argmin}

\newcommand{\interior}{\operatorname{int}}

%% file: tables/concave_table.tex
\setlength{\intextsep}{0pt}
\begin{wraptable}[15]{r}{0.5\textwidth}
\vspace{-4ex}
\centering
\small
\caption{Solutions to \cref{ex:concave} recovered by the Lagrangian approach with PI multiplier updates. At every constraint level tested, the method recovers the constrained minimizer $x^* = \arcsin(\epsilon)$.}
\label{table:concave}
\vspace{1ex}
\begin{tabular}{ccc}
\toprule
$\epsilon$ & $x^* = \arcsin(\epsilon)$ & $x_t$ at convergence \\
\midrule
$0.1$ & $1.00 \times 10^{-1}$ & $1.00 \times 10^{-1}$ \\
$0.2$ & $2.01 \times 10^{-1}$ & $2.01 \times 10^{-1}$ \\
$0.3$ & $3.05 \times 10^{-1}$ & $3.05 \times 10^{-1}$ \\
$0.4$ & $4.12 \times 10^{-1}$ & $4.12 \times 10^{-1}$ \\
$0.5$ & $5.24 \times 10^{-1}$ & $5.24 \times 10^{-1}$ \\
$0.6$ & $6.44 \times 10^{-1}$ & $6.44 \times 10^{-1}$ \\
$0.7$ & $7.75 \times 10^{-1}$ & $7.75 \times 10^{-1}$ \\
$0.8$ & $9.27 \times 10^{-1}$ & $9.27 \times 10^{-1}$ \\
\bottomrule
\end{tabular}
\end{wraptable}
\setlength{\intextsep}{12pt plus 2pt minus 2pt}

%% file: appendix.tex
\onecolumn

\appendixwithtoc

\newpage

\section{Proofs}
\label{app:proofs}

\begin{proof}[\textbf{Proof of Proposition} \ref{prop:penalized}]
    \hypertarget{proof:penalized}{}
    Since \( \xstar \) is an interior constrained minimizer with corresponding Lagrange multipliers \( \lambdastar \vgeq \vzero \) and \( \mustar \), the KKT stationarity condition gives
    \begin{equation}
        \nabla f(\xstar) + \vlambda^{*\top} \nabla \vg(\xstar) + \vmu^{*\top} \nabla \vh(\xstar) = \vzero .
    \end{equation}
    Choose \( \vcg = \lambdastar \) and \( \vch = \mustar \). The stationarity condition then implies that \( \xstar \) is a critical point of the penalized objective
    \[
        \P(\vx) = f(\vx) + \vcg^\top \vg(\vx) + \vch^\top \vh(\vx).
    \]
    The objective \( \P \) is convex in \( \vx \): \( f \) is convex, \( \vcg^\top \vg \) is convex because \( \vcg = \lambdastar \vgeq \vzero \), and \( \vch^\top \vh \) is affine. For convex functions, any critical point is a global minimizer. Hence,
    \[
        \xstar \in \argmin_{\vx \in \X} \P(\vx).
    \]
\end{proof}

\medskip

\begin{proof}[\textbf{Proof of Proposition} \ref{prop:solution}]
    \hypertarget{proof:solution}{}
    Since $\epsilon \in (0,1)$ and $\sin(x)$ is strictly increasing on $[0,\pi/2]$, the feasibility condition
    \[
        g(x)=\sin(x)\le \epsilon
    \]
    is equivalent to
    \[
        x \in [0,\arcsin(\epsilon)].
    \]
    Thus, \cref{ex:concave} reduces to
    \[
        \min_{x \in [0,\arcsin(\epsilon)]} \cos(x).
    \]

    Because $\cos(x)$ is strictly decreasing on $[0,\pi/2]$, its minimum over the feasible interval is attained at the largest feasible point, namely
    \[
        x^*=\arcsin(\epsilon).
    \]
    At this point,
    \[
        g(x^*)=\sin(\arcsin(\epsilon))=\epsilon,
        \qquad
        f(x^*)=\cos(\arcsin(\epsilon))=\sqrt{1-\epsilon^2}.
    \]
    Hence the constraint is active at the constrained minimizer.

    It remains to identify the corresponding Lagrange multiplier. The Lagrangian is
    \[
        \Lag(x,\lambda)=\cos(x)+\lambda(\sin(x)-\epsilon),
        \qquad \lambda\ge 0.
    \]
    Complementary slackness holds automatically because the constraint is active at $x^*$. Since $x^*$ lies in the interior of $[0,\pi/2]$, stationarity requires
    \[
        \frac{\partial \Lag}{\partial x}(x^*,\lambda^*)
        =
        -\sin(x^*)+\lambda^*\cos(x^*)
        =0.
    \]
    Substituting $x^*=\arcsin(\epsilon)$ gives
    \[
        -\epsilon+\lambda^*\sqrt{1-\epsilon^2}=0,
    \]
    and therefore
    \[
        \lambda^*=\frac{\epsilon}{\sqrt{1-\epsilon^2}}.
    \]
    This proves the claim.
\end{proof}

\medskip

\begin{proof}[\textbf{Proof of Proposition} \ref{prop:concave}]
    \hypertarget{proof:concave}{}
    Consider
    \[
        \P_c(x)=\cos(x)+c\sin(x),
        \qquad x\in[0,\pi/2], \quad c\ge 0.
    \]
    Its second derivative is
    \[
        \P_c''(x) = -\cos(x)-c\sin(x).
    \]
    Since $\cos(x)>0$ on $[0,\pi/2)$ and $\sin(x)\ge 0$ on $[0,\pi/2]$, we have $\P_c''(x)<0$ on $[0,\pi/2)$ for every $c\ge 0$, so $\P_c$ is strictly concave on $[0,\pi/2]$. Its minimum over this interval is therefore attained only at an endpoint.

    Evaluating the endpoints gives
    \[
        \P_c(0)=1,
        \qquad
        \P_c(\pi/2)=c.
    \]
    Therefore:
    \begin{itemize}
        \item if $c<1$, then $\P_c(\pi/2)=c<1=\P_c(0)$, so the unique minimizer is $x_c=\pi/2$;
        \item if $c>1$, then $\P_c(0)=1<c=\P_c(\pi/2)$, so the unique minimizer is $x_c=0$;
        \item if $c=1$, then $\P_c(0)=\P_c(\pi/2)=1$, so both $x_c=0$ and $x_c=\pi/2$ are minimizers.
    \end{itemize}

    Finally, evaluating the objective and constraint at these points,
    \[
        f(0)=1,\qquad g(0)=0\le \epsilon,
    \]
    and
    \[
        f(\pi/2)=0,\qquad g(\pi/2)=1>\epsilon.
    \]
    This proves the characterization of the minimizers and their feasibility properties.
\end{proof}

\section{Experimental Details}

Our implementations use PyTorch~\citep{pytorch} and the Cooper library for constrained deep learning~\citep{gallegoPosada2025cooper}. 
Our code is available at: \texttt{\url{https://github.com/merajhashemi/constraints-vs-penalties}}.

\subsection{Example 1}
\label{app:concave}

\begin{figure}[ht]
    \centering
    \includegraphics[width=0.5\linewidth]{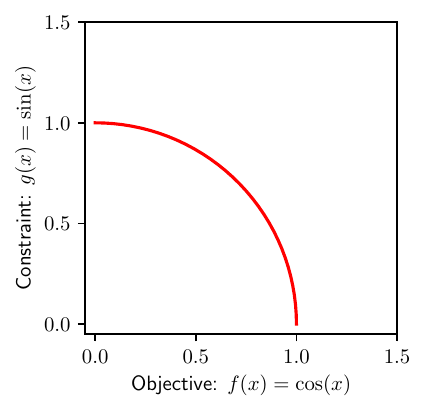}
    \caption{
        The trade-off curve between the objective \(f\) and constraint \(g\) in \cref{ex:concave}. As \(x\) varies over \([0,\pi/2]\), the map \(x \mapsto (f(x), g(x)) = (\cos x, \sin x)\) traces the non-dominated frontier from \((1,0)\) to \((0,1)\). The constrained optimum is the interior point \((\sqrt{1-\epsilon^2},\,\epsilon)\), whereas fixed weighted-sum penalization recovers only the two endpoints.
    }
    \label{fig:pareto}
\end{figure}

\Cref{fig:pareto} illustrates the trade-off between the objective and constraint values in \cref{ex:concave}. 
Imposing the constraint \(g(x) \leq \epsilon\) restricts attention to the portion of the trade-off curve on which \(g\) is at most \(\epsilon\). Since minimizing \(f(x)=\cos x\) over \([0,\pi/2]\) amounts to increasing \(x\), the constrained minimizer is the largest feasible point, where \(g(x)=\epsilon\), namely \(x^*=\arcsin(\epsilon)\).

Intuitively, the concavity of this trade-off curve—closely related to the concavity of the penalized objective in this example—causes gradient descent on the penalized formulation to converge to the two endpoints of the curve. By contrast, under the convexity assumptions of \cref{prop:penalized}, varying the penalty coefficient would also recover intermediate trade-offs.

See \citet{blog_1, blog_2} for animations illustrating the optimization dynamics of \blobletter{1} gradient descent on the penalized formulation, which converges to the endpoints; \blobletter{2} gradient descent--ascent on the Lagrangian, which oscillates without converging; and \blobletter{3} gradient descent--ascent on the Augmented Lagrangian, which does converge.

\textbf{Hyperparameters}.
We solve \cref{ex:concave} with the Lagrangian approach. The primal optimizer is gradient descent with a step-size of \(0.01\). The dual optimizer is $\nu$PI~\citep{sohrabi2024nupi}, as implemented in Cooper~\citep{gallegoPosada2025cooper}, with a step-size of \(0.3\), damping coefficient \(K_p = 40\), and \(\nu = 0\), which reduces $\nu$PI to a PI controller. The primal and dual variables are updated simultaneously. Cooper scales \(K_p\) by the step-size, so this damping coefficient is \(\kappa_p = 12\) in the notation of \citet{sohrabi2024nupi}.
The variable is initialized at $x = 0.5$ and the multiplier at zero, and $x$ is projected onto \([0, \pi/2]\) after every primal update.
\Cref{table:concave} reports $x$ after 1,000 iterations; at every constraint level, the multiplier also reaches $\lambda^* = \epsilon/\sqrt{1-\epsilon^2}$. A damped dual update is necessary for convergence here. With plain gradient ascent ($K_p = 0$), the iterates do not converge: depending on $\epsilon$, they oscillate around the solution~\citep{platt1988constrained} or stop at an endpoint of \([0, \pi/2]\), where the gradient of \(f\) (at \(0\)) or of \(g\) (at \(\pi/2\)) vanishes.

The Lagrangian of \cref{ex:concave} is concave in \(x\) for every multiplier, so even at the optimal multiplier, \(x^*\) is a stationary point of the Lagrangian in \(x\) but not a minimizer. The Augmented Lagrangian method~\citep{hestenes1969multiplier,powell1969method,rockafellar1973dual} adds a quadratic penalty on constraint violations to the Lagrangian and, for a large enough penalty parameter, makes a constrained minimizer satisfying LICQ and the second-order sufficient conditions a strict local minimizer in \(x\) at the optimal multiplier~\citep[Thm.~17.5]{nocedal2006numerical}. PI multiplier updates are equivalent to the Augmented Lagrangian method~\citep{ramirez2025dual}, which explains why the damped updates converge here.

\subsection{Sparsity Constraints}
\label{app:sparsity}

\citet{louizos2017learning} propose a model reparameterization that enables $L_0$-norm regularization via stochastic gates $\vz \in [0,1]^d$, one per weight, following a Hard Concrete distribution with parameters $\vphi$. Their penalized formulation is
\begin{equation}
    \label{eq:sparsity_cmp_app}
    \min_{\vw, \vphi \in \mathbb{R}^d}
    \mathbb{E}_{\vz | \vphi} \left[ L(\vw \odot \vz \,|\, \mathcal{D}) \right]
    + c \cdot \mathbb{E}_{\vz | \vphi} [\|\vz\|_0],
\end{equation}
where $\vw$ are model parameters, $L$ the task loss, $\mathcal{D}$ the dataset, $\|\cdot\|_0$ the $L_0$ norm, and $c > 0$ a penalty coefficient. In our implementation, the gates are instead structured, one per input unit of each layer, so that a gate masks all of that unit's outgoing weights, and the penalty $\mathbb{E}_{\vz | \vphi} [\|\vz\|_0]$ is replaced by the expected model density, the expected fraction of non-bias weights that are active. The coefficient $c$ in \cref{fig:sparsity} and \cref{table:sparse} multiplies this density.

To illustrate the tuning burden of the penalized approach versus constrained methods, we replicate the bisection search experiment from Figure~5(b), Appendix~E of \citet{gallego2022controlled}. We use their MNIST classification setup with an MLP (300–100 hidden units), as shown in \cref{fig:sparsity}. To meet the constraint of at most 50\% model density, we perform a log-scale bisection search over the penalty coefficient $c \in [10^{-3}, 1]$, stopping once the model density is within one percentage point of 50\%. The corresponding results, including training accuracy, are reported in \cref{table:sparse} in Appendix \ref{app:sparsity_results}.

\textbf{Hyperparameters}. Following \citet{gallego2022controlled}, we set the temperature of the gate distribution to $2/3$ and use a stretching interval of $[-0.1, 1.1]$. The initial drop rate is $0.2$ for the input layer and $0.5$ for the other layers. We use Adam~\citep{kingma2015adam} with a step-size of $0.001$ and no weight decay to optimize both the model and gate parameters, and train for $150$ epochs with a batch size of $256$. Density and accuracy are measured on the training set with the deterministic test-time gates.

\subsection{Probability Constraints}
\label{app:rate}

We consider a linear binary classification problem in which the average predicted probability of class~0, $P(\hat{y} = 0) = \mathbb{E}_{(x, y) \sim \mathcal{D}} \left[1 - \sigma(w^\top x + b)\right]$, must be at least 70\%. The problem is
\begin{equation}
\label{eq:rate}
\min_{w, b} \ \mathbb{E}_{(x, y) \sim \mathcal{D}} \left[ \ell(w^\top x + b, y) \right] \quad \text{s.t.} \quad P(\hat{y} = 0) \geq 0.7,
\end{equation}
where $\ell$ denotes the cross-entropy loss, $w$ and $b$ are the weights and bias of the linear model, $(x, y)$ are input-label pairs drawn from the data distribution $\mathcal{D}$, and $\sigma$ is the sigmoid function. The data and linear model are those of Figure~2 of \citet{cotter2019proxy}, whose constraint is instead on the rate of hard predictions and is handled with proxy constraints. We constrain the average predicted probability, which is differentiable, so that the penalized and Lagrangian approaches solve the same problem and differ only in how the coefficient is set.

The penalized formulation of \cref{eq:rate} adds the constraint function $0.7 - P(\hat{y} = 0)$ to the objective with a fixed coefficient, which up to an additive constant gives
\begin{equation}
\label{eq:rate_pen}
\min_{w, b} \ \mathbb{E}_{(x, y) \sim \mathcal{D}} \left[ \ell(w^\top x + b, y) \right] - c \cdot P(\hat{y} = 0),
\end{equation}
where $c>0$ is a penalty coefficient.

In \Cref{fig:robustness}, we present a sweep over the dual step-size when solving \cref{eq:rate} with the Lagrangian approach, and over the penalty coefficient when solving \cref{eq:rate_pen}. The results illustrate that tuning dual step-sizes in the Lagrangian approach is much easier than tuning penalty coefficients. The same results, including the training accuracy of each model and the final value of the multiplier, are provided in \cref{table:surrogate_rate_constraint} in Appendix \ref{app:rate_results}.

\textbf{Hyperparameters}. For the constrained approach, we use gradient descent--ascent with a primal step-size of $2 \times 10^{-2}$, training for $10{,}000$ iterations. The dual step-size is swept over several orders of magnitude. The model is initialized at $w = (1, -1)$ and $b = 0$, and the multiplier at zero. For the penalized approach, we use the same primal optimization pipeline—gradient descent with a step-size of $2 \times 10^{-2}$—and sweep over penalty coefficients using the same set of values as for the dual step-size. The data are $400$ linearly separable points in two dimensions, drawn from a mixture of four Gaussians, two per class, with $100$ points each. Training is done using full-batch optimization.

\section{Comprehensive Experimental Results}
\label{app:results}

This appendix provides the full tabular results complementing the experiments in the main paper: \cref{fig:sparsity} (\S\ref{sec:tuning}) and \cref{fig:robustness} (\S\ref{sec:contrasting}).

\subsection{Sparsity Constraints}
\label{app:sparsity_results}

\Cref{table:sparse} reports the same results as the bisection-search experiment in \Cref{fig:sparsity}, now also including the training accuracy of each model at convergence. Task and experimental details are provided in Appendix \ref{app:sparsity}. As the results show, overshooting the constraint can hurt performance by unnecessarily reducing model capacity. For example, with $c = 1$, the method returns a feasible solution with $25.3\%$ model density ($74.7\%$ sparsity), but this comes at a performance cost. The solution should instead lie closer to the constraint boundary, as in the final coefficient choice of $2.21 \times 10^{-1}$.

\input{tables/sparse_table}

\subsection{Probability Constraints}
\label{app:rate_results}

\Cref{table:surrogate_rate_constraint} reports the same results as the probability-constrained classification experiment in \Cref{fig:robustness}, now also including the training accuracy of each model at convergence and the final value of the multiplier. The task and experimental setup are described in Appendix \ref{app:rate}. Most penalty coefficients lead to collapsed solutions: either the model effectively optimizes only for accuracy and keeps the average predicted probability of class 0 near $50\%$, or it focuses almost entirely on the penalty and pushes that probability towards $100\%$. As in the sparsity-constrained experiment (\cref{table:sparse}), overshooting the constraint harms performance by placing excessive weight on the penalty, which conflicts with the objective. In contrast, the Lagrangian approach recovers the desired solution for most of the dual step-sizes considered.

\input{tables/rate_constraints_surrogate_table}

%% file: tables/sparse_table.tex
\begin{table}[H]

\centering
\caption{Sparsity-constrained neural network training using the penalized approach, targeting at most 50\% model density (at least 50\% sparsity), within one percentage point. The penalty coefficient is selected via a log-scale bisection search. This table complements \Cref{fig:sparsity}.}
\label{table:sparse}
\begin{tabular}{c|lcr}
\toprule
Iteration \# & Penalty coef. $c$ & Model density (\%) & Train acc. (\%) \\
\midrule
0 & $1.00 \times 10^{-3}$ & $84.5$ & $99.97$ \\
0 & $1.00 \times 10^{0}$ & $25.3$ & $99.97$ \\
1 & $3.16 \times 10^{-2}$ & $68.1$ & $99.96$ \\
2 & $1.78 \times 10^{-1}$ & $54.2$ & $100.00$ \\
3 & $4.22 \times 10^{-1}$ & $39.0$ & $100.00$ \\
4 & $2.74 \times 10^{-1}$ & $46.9$ & $99.99$ \\
5 & $2.21 \times 10^{-1}$ & $50.5$ & $99.99$ \\
\bottomrule
\end{tabular}

\vspace{-1ex}
\end{table}

%% file: tables/rate_constraints_surrogate_table.tex
\begin{table}[H]
\centering
\small
\caption{Probability-constrained linear classification (Appendix~\ref{app:rate}), targeting an average predicted probability of class 0, $P(\hat{y}=0)$, of at least $70\%$. Both sweeps use the same grid of values, as penalty coefficients $c$ for the penalized approach and as dual step-sizes for the Lagrangian approach. The multiplier at the end of training is $\lambda_T$. The Lagrangian approach lands on the target for 15 of the 20 dual step-sizes, whereas only $c = 2.15$ avoids a collapsed penalized solution, overshooting to $75\%$.}
\label{table:surrogate_rate_constraint}
\begin{tabular}{c|cc|ccc}
\toprule
& \multicolumn{2}{c|}{Penalized ($c$)} & \multicolumn{3}{c}{Lagrangian (dual step-size)} \\
Value & $P(\hat{y}=0)$ (\%) & Train acc. (\%) & $P(\hat{y}=0)$ (\%) & Train acc. (\%) & $\lambda_T$ \\
\midrule
$0$ & $49.96$ & $99.75$ & $49.96$ & $99.75$ & $0.00$ \\
$1.00 \times 10^{-4}$ & $49.96$ & $99.75$ & $50.30$ & $99.75$ & $0.20$ \\
$2.15 \times 10^{-4}$ & $49.96$ & $99.75$ & $50.72$ & $99.75$ & $0.42$ \\
$4.60 \times 10^{-4}$ & $49.96$ & $99.75$ & $51.89$ & $100.00$ & $0.87$ \\
$1.00 \times 10^{-3}$ & $49.96$ & $99.75$ & $59.56$ & $96.25$ & $1.63$ \\
$2.15 \times 10^{-3}$ & $49.96$ & $99.75$ & $69.96$ & $84.25$ & $1.86$ \\
$4.60 \times 10^{-3}$ & $49.97$ & $99.75$ & $69.93$ & $84.25$ & $1.83$ \\
$1.00 \times 10^{-2}$ & $49.98$ & $99.75$ & $69.96$ & $84.25$ & $1.83$ \\
$2.15 \times 10^{-2}$ & $50.00$ & $99.75$ & $69.98$ & $84.25$ & $1.83$ \\
$4.60 \times 10^{-2}$ & $50.06$ & $99.75$ & $69.99$ & $84.25$ & $1.83$ \\
$1.00 \times 10^{-1}$ & $50.17$ & $99.75$ & $70.00$ & $84.25$ & $1.83$ \\
$2.15 \times 10^{-1}$ & $50.43$ & $99.75$ & $70.00$ & $84.25$ & $1.83$ \\
$4.60 \times 10^{-1}$ & $51.08$ & $99.75$ & $70.00$ & $84.25$ & $1.83$ \\
$1.00 \times 10^{0}$ & $54.24$ & $99.50$ & $70.00$ & $84.25$ & $1.83$ \\
$2.15 \times 10^{0}$ & $74.98$ & $74.00$ & $70.00$ & $84.25$ & $1.83$ \\
$4.60 \times 10^{0}$ & $88.89$ & $50.25$ & $70.00$ & $84.25$ & $1.83$ \\
$1.00 \times 10^{1}$ & $94.96$ & $50.00$ & $70.00$ & $84.25$ & $1.83$ \\
$2.15 \times 10^{1}$ & $97.67$ & $50.00$ & $70.00$ & $84.25$ & $1.83$ \\
$4.60 \times 10^{1}$ & $98.91$ & $50.00$ & $70.00$ & $84.25$ & $1.83$ \\
$1.00 \times 10^{2}$ & $99.50$ & $50.00$ & $70.00$ & $84.25$ & $1.83$ \\
\bottomrule
\end{tabular}
\end{table}